\documentclass{article}

\usepackage{microtype}
\usepackage{graphicx}
\usepackage{booktabs} %

\usepackage{hyperref}

\usepackage{amsmath}
\usepackage{amssymb}
\usepackage{mathtools}
\usepackage{amsthm}
\usepackage{multirow}
\usepackage{subcaption}
\usepackage{caption}
\usepackage{wrapfig}
\usepackage[inline]{enumitem}

\usepackage[accepted]{icml2024}
\usepackage{breakurl}

\usepackage[capitalize,noabbrev]{cleveref}

\theoremstyle{plain}

\theoremstyle{definition}

\theoremstyle{remark}

\usepackage[textsize=tiny]{todonotes}
\usepackage{nicefrac}
\usepackage{array}

\usepackage{tcolorbox}
\tcbuselibrary{listings}
\newtcblisting{inlinecode}{
  listing only,
  nobeforeafter,
  after={\xspace},
  hbox,
  tcbox raise base,
  fontupper=\ttfamily,
  colback=lightgray,
  colframe=lightgray,
  size=fbox
  }

\icmltitlerunning{Accelerating Simulation of Two-Phase Flows with Neural PDE Surrogates}

\begin{document}

\twocolumn[
\icmltitle{Accelerating Simulation of Two-Phase Flows with Neural PDE Surrogates}

\icmlsetsymbol{equal}{*}

\begin{icmlauthorlist}
\icmlauthor{Yoeri Poels}{equal,spc,dai}
\icmlauthor{Koen Minartz}{equal,dai}
\icmlauthor{Harshit Bansal}{equal,casa}
\icmlauthor{Vlado Menkovski}{dai}
\end{icmlauthorlist}

\icmlaffiliation{spc}{Swiss Plasma Center, École Polytechnique Fédérale de Lausanne}
\icmlaffiliation{dai}{Data and Artificial Intelligence Cluster, Eindhoven University of Technology}
\icmlaffiliation{casa}{Centre for Analysis, Scientific Computing, and Applications, Eindhoven University of Technology}

\icmlcorrespondingauthor{Yoeri Poels}{yoeri.poels@epfl.ch}
\icmlcorrespondingauthor{Koen Minartz}{k.minartz@tue.nl}
\icmlcorrespondingauthor{Harshit Bansal}{h.bansal@tue.nl}
\icmlcorrespondingauthor{Vlado Menkovski}{v.menkovski@tue.nl}

\icmlkeywords{Neural PDE surrogates, Neural PDE solvers, two-phase flows, multiphase flows, multiphysics problems, oil expulsion}

\vskip 0.3in
]

\printAffiliationsAndNotice{\icmlEqualContribution} %
\begin{abstract}
Simulation is a powerful tool to better understand physical systems, but generally requires computationally expensive numerical methods. Downstream applications of such simulations can become computationally infeasible if they require many forward solves, for example in the case of inverse design with many degrees of freedom. In this work, we investigate and extend neural PDE solvers as a tool to aid in scaling simulations for two-phase flow problems, and simulations of oil expulsion from a pore specifically. We extend existing numerical methods for this problem to a more complex setting involving varying geometries of the domain to generate a challenging dataset. Further, we investigate three prominent neural PDE solver methods, namely the UNet, DRN, and U-FNO, and extend them for characteristics of the oil-expulsion problem: (1) spatial conditioning on the geometry; (2) periodicity in the boundary; (3) approximate mass conservation. We scale all methods and benchmark their speed-accuracy trade-off, evaluate qualitative properties, and perform an ablation study. We find that the investigated methods can accurately model the droplet dynamics with up to three orders of magnitude speed-up, that our extensions improve performance over the baselines, and that the introduced varying geometries constitute a significantly more challenging setting over the previously considered oil expulsion problem.

\end{abstract}

\section{Introduction}
\label{sec:introduction}

\begin{figure*}[t]
\centering
\hfill
\raisebox{2.45cm}{\rotatebox{90}{\textbf{Reference}}}\hspace{0.13cm}
\includegraphics[width=.18\linewidth]{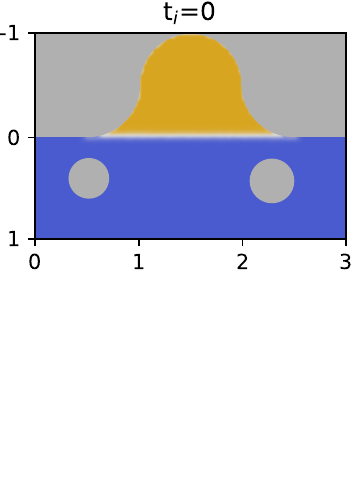}\hfill
\hspace{-0.33cm}\raisebox{0.4cm}{\rotatebox{90}{\textbf{Prediction}}}\hspace{0.13cm}
\includegraphics[width=.18\linewidth]{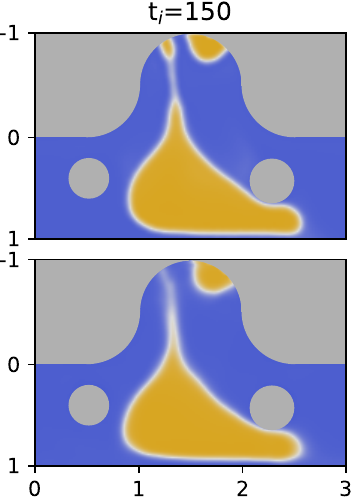}\hfill
\includegraphics[width=.18\linewidth]{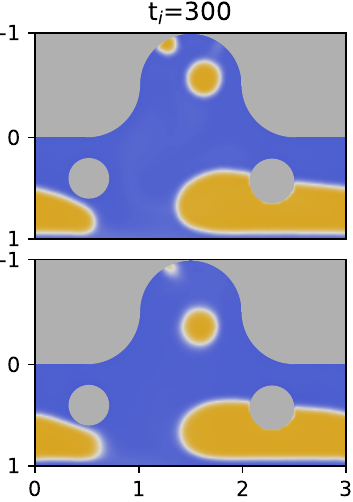}\hfill
\includegraphics[width=.18\linewidth]{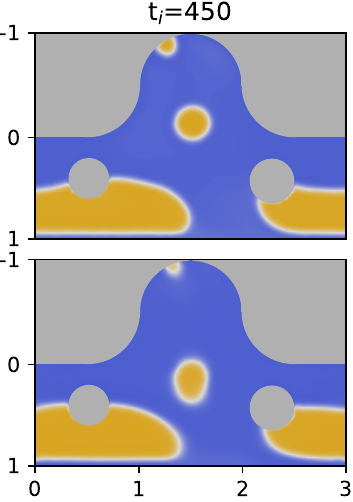}\hfill
\hfill%
\centering
\caption{
\centering
\begin{minipage}{0.915\textwidth}
\centering
\makebox[0pt]{The oil expulsion problem illustrated. \textit{Top}: Numerical solver reference. \textit{Bottom}: Neural surrogate model (UNet) prediction.} %
\end{minipage}}
\label{fig:problem}
\end{figure*}

    Multiphysics phenomena emerge from the interplay of different physical aspects in a system. Such phenomena appear naturally in various domains, such as digital microfluidics and reservoir engineering, among many others~\citep{microfluidics,linga2019bernaise}. In this work, we focus on a particular type of multiphysics process, namely multiphase electrohydrodynamics~\citep{linga2019bernaise}. %
    Specifically, we investigate the industrially relevant application of oil expulsion from a pore. %
    A two-phase flow of water and oil is modeled, where electrowetting is applied to an oil droplet to change the droplet's properties using an electrical charge. An illustration of this process can be found in Figure~\ref{fig:problem}.

    Some of the key challenges in this application are to control the spatiotemporal evolution of the dynamics, alongside the coalescence and break-down of the oil droplets. Generally we desire a specific dynamic response, which makes configuring operating conditions like electrical charge and rate of flow in the presence of varying geometries a challenging task. It is essentially an ill-posed inverse problem, requiring several forward solves to find a good solution. However, solving a high-fidelity forward model requires substantial computational resources, making it unpractical for multi-query optimization or real-time control.

     Moreover, while extensive studies have been done to obtain stable, accurate, robust, and efficient high-fidelity numerical methods for simulating the aforementioned phenomena~\citep{linga2019bernaise,JCPlinga, ALMASI2021}, these methods were deployed in a setting without geometrical obstacles. To create a more challenging setting, we adapt these methods for obstacles in the domain. However, this extension leads to a further increase in computational cost, making the need for fast approximate methods significant.

    To alleviate the issue of computationally expensive simulations, we propose the use of neural PDE solvers for fast surrogate modeling. We employ these models in an autoregressive manner by approximating the time-stepping operator of the system. While constructing approximate solutions of PDEs with neural networks has gained considerable traction~\cite{azizzadenesheli2024}, there is no comprehensive study of neural PDE surrogates applied to two-phase flows modeling oil extraction. In this work, we conduct such an investigation. In summary, our contributions are as follows:

    \begin{itemize}
        \item We extend the numerical method for the oil expulsion problem as introduced in~\cite{linga2019bernaise} to account for geometrical obstacles in the spatial domain;
        \item We generate a dataset of approximately 1000 simulations, alongside a simplified dataset without geometrical obstacles of approximately 500 simulations as a baseline setting, following~\citet{linga2019bernaise};
        \item We extend three prevalent neural PDE surrogate modeling methods~\cite{stachenfeld2021learned,gupta2022pdearena,wen2022ufno} to account for the characteristics specific to the oil expulsion problem;
        \item We provide a benchmark of these models, evaluating their scaling in terms of accuracy and inference speed, and provide a sensitivity analysis comparing the surrogate models with the baseline high-fidelity solver.        
    \end{itemize}

\section{Background and related work}
\label{sec:related-work}

\subsection{Oil expulsion problem}

The oil expulsion problem is modeled as a multiphase electrohydrodynamical system. Specifically, it is governed by the following system of coupled PDEs:
\begin{equation}
\begin{aligned}\label{eq1}
    &\partial_t(\rho(\phi) \mathbf{v})+\nabla \cdot(\rho(\phi) \mathbf{v} \otimes \mathbf{v})\\
    &-\nabla \cdot[2 \mu(\phi) \mathcal{D} \mathbf{v} 
+\mathbf{v} \otimes \rho^{\prime}(\phi) M(\phi) \nabla {g_\phi}]+\nabla p\\
=&-\phi \nabla {g_\phi}-\sum_i c_i \nabla {g_{c_i}},
\end{aligned}
\end{equation}\vspace{-.6cm}
\begin{align}
    \nabla \cdot \mathbf{v}&=0, \label{eq2}\\
\partial_t \phi+\mathbf{v} \cdot \nabla \phi-\nabla \cdot\left(M(\phi) \nabla {g_\phi}\right)&=0, \label{eq3}\\
\partial_t c_j+\mathbf{v} \cdot \nabla {c_j}-\nabla \cdot\left(K_j(\phi) c_j \nabla {g_{c_j}}\right)&=0, \label{eq4}\\
\nabla \cdot(\varepsilon(\phi) \nabla V)&=-\rho_e, \label{eq5}
\end{align}
where $t$ refers to time, $\mathbf{v}$ to the velocity field, $p$ to the pressure field, $\phi$ to the phase field, $\rho$ to the density ($\rho_e$ the charge density), $\mu$ to the viscosity, $c_j$ to the solute concentration of species $j \in \{\text{H}^\pm\}$, $g_{c_j}$ to the electrochemical potential of a species, $g_{\phi}$ to the electrochemical potential of the phase field, $K_j$ to the species' diffusivities, $M$ to the phase field mobility, $V$ to the electrostatic potential, $\varepsilon$ to the permissivity, and $\mathcal{D}\mathbf{v}$ to the symmetric velocity gradient.
Equation~\ref{eq1} describes the momentum balance, Equation~\ref{eq2} the continuity equation, Equation~\ref{eq3} the conservative evolution of the phase field, Equation~\ref{eq4} governs the transport of the concentration field of species, and Equation~\ref{eq5} describes Gauss's law.

To obtain high-fidelity solutions, 
we employ a linear operator splitting scheme~\cite{linga2019bernaise} that results in three different subproblems - \emph{(i)} hydrodynamic flow, \emph{(ii)} phase field transport, and \emph{(iii)} chemical transport under an electric field. %
We build on the implementation in BERNAISE~\cite{linga2019bernaise}, a FEniCS-based~\cite{fenics2015} high-fidelity solver, and adapt it to account for multiple geometrical obstacles and different practically relevant boundary conditions. For more details on the initial conditions, boundary conditions and the employed numerical method, we refer to~\citet{linga2019bernaise}, to which we added boundaries around the obstacles. In particular, we have a non-zero Neumann condition for the electric potential, and a no-slip condition for the velocity field.

\subsection{Neural PDE surrogate modeling}

Constructing approximate solutions of PDEs with neural networks has seen a surge in interest over the past few years, with example applications in fluid dynamics~\citep{li2021fno, brandstetter2022message, gupta2022pdearena}, climate and weather modeling~\citep{nguyen2023climax, bonev2023sfno, gao2023prediff}, and thermonuclear fusion~\citep{Poels2023neuralPDE, Gopakumar2024PlasmaSurrogate}, among others. However, compared to these settings, the simulation of two-phase flow problems with neural networks that we consider is relatively underexplored. 

One relatively popular used approach for solving two-phase flow problems with neural networks is Physics-Informed Neural Networks (PINNs)~\citep{raissi2019pinn}. 
In this context, example applications of PINNs include the simulation of two-phase flows in porous media~\citep{Zhang2023Darcy, Hanna2022Porous}, simulation of the interface of two fluids~\citep{Zhu2023Interface}, simulating two-phase material microstructures~\citep{Ren2024twophasemat}, and the simulation of the dynamics of bubbles~\citep{Lin2021Bubble,zhai2022BubblePinn,Buhenda2021Infer}. Although PINNs inherently exploit domain knowledge by minimizing the residual of the PDE directly, they generally need to be optimized from scratch for every PDE instance and parameterization. As one of the main goals of our work is accelerating two-phase flow simulations, this makes them impractical for our setting.

Consequently, we focus on a different approach, namely autoregressive models and neural operators. We will refer to these methods as neural surrogate models. Compared to PINNs, neural surrogate models typically cannot exploit domain knowledge as easily, and rely more on the availability of data. However, once trained, they can be used to get approximate solutions to multiple parameterizations of the system of PDEs in question. There have been various applications of popular neural surrogate architectures to two-phase flow and multiphase flow problems, for example Deep-O-Nets~\citep{Lu2024DeepONet, Lin2021DeepOBubble}, Fourier Neural Operators (FNOs)~\citep{Hassan2023BubbleML, jiang2023fouriermionet}, and convolutional architectures~\citep{Hassan2023BubbleML}. However, none of these works provide a comprehensive, independent and consistent comparison of neural PDE architectures in this setting. %

\section{Method}\label{sec:method}

\begin{figure*}[t]
\centering
\begin{subfigure}[t]{0.57\linewidth}
\centering
\includegraphics[width=\linewidth]{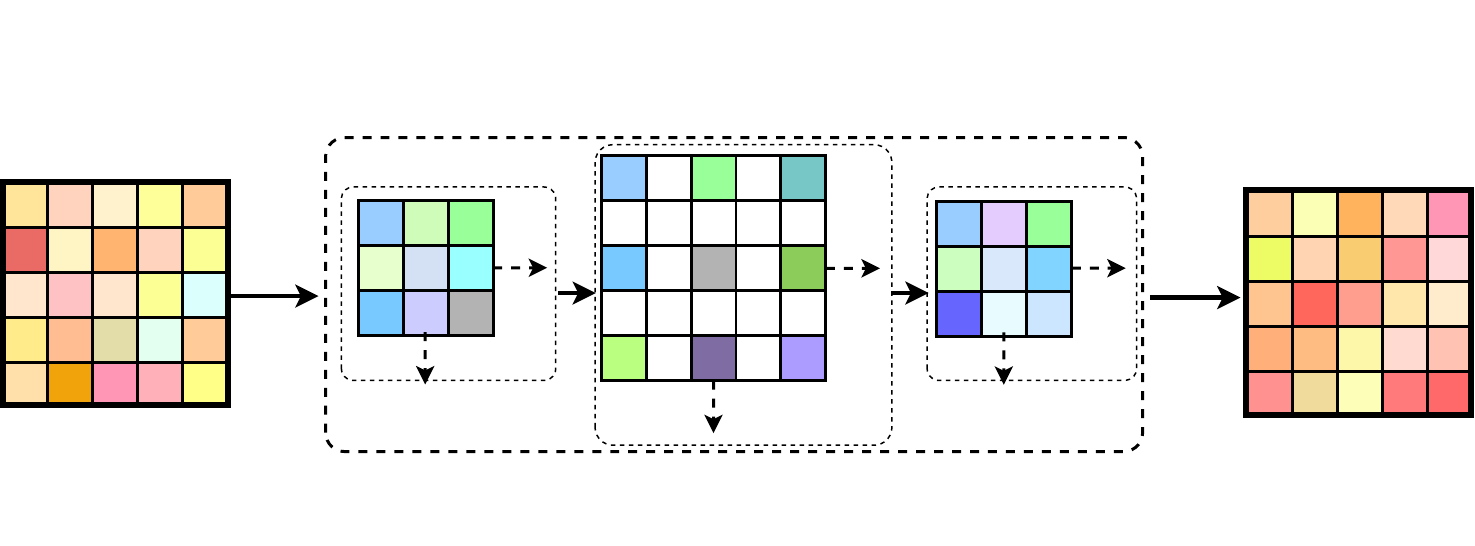}
\caption{Dilated ResNet~\cite{stachenfeld2021learned}.}\label{subfig:DRN-diagram}
\end{subfigure}
\hfill
\begin{subfigure}[t]{0.415\linewidth}
\centering
\includegraphics[width=\linewidth]{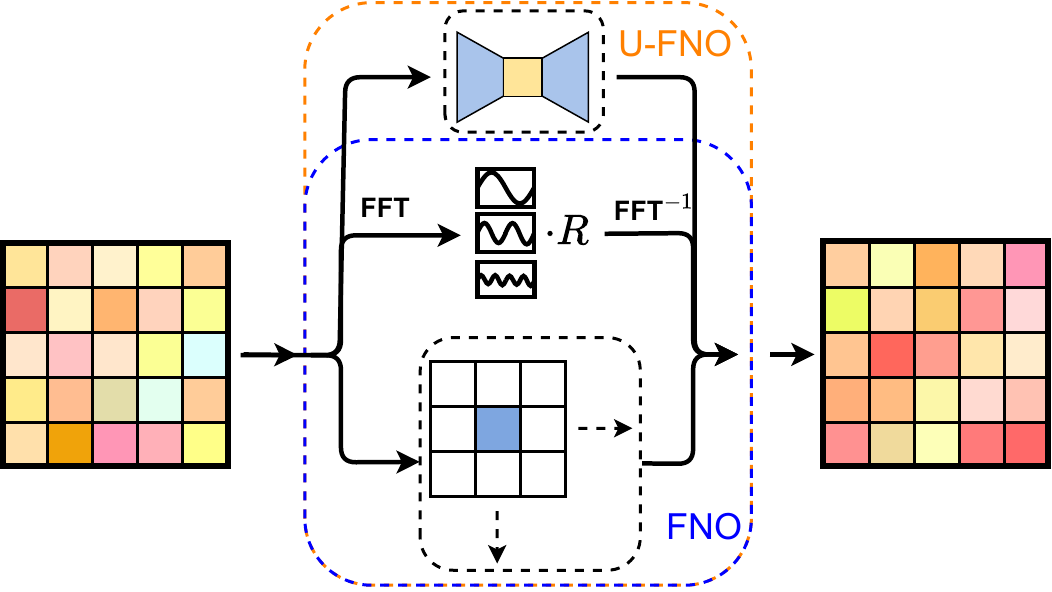}
\caption{Fourier Neural Operator and U-FNO~\citep{li2021fno, wen2022ufno}.}
\label{subfig:FNO-diagram}
\end{subfigure}\\
\vspace{2mm}
\begin{subfigure}[t]{0.82\linewidth}
\centering
\includegraphics[width=\linewidth]{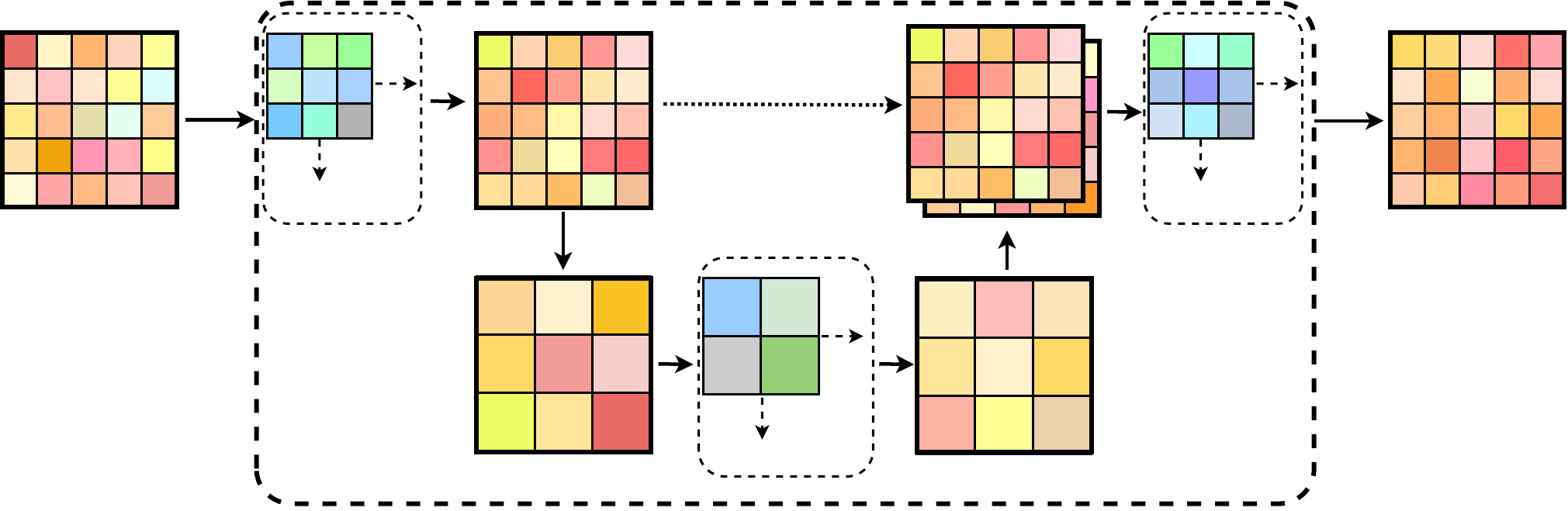}
\caption{UNet~\cite{ronneberger2015, gupta2022pdearena}.}
\label{subfig:UNet-diagram}
\end{subfigure}
\hfill
\begin{subfigure}[t]{0.17\linewidth}
\includegraphics[width=\linewidth]{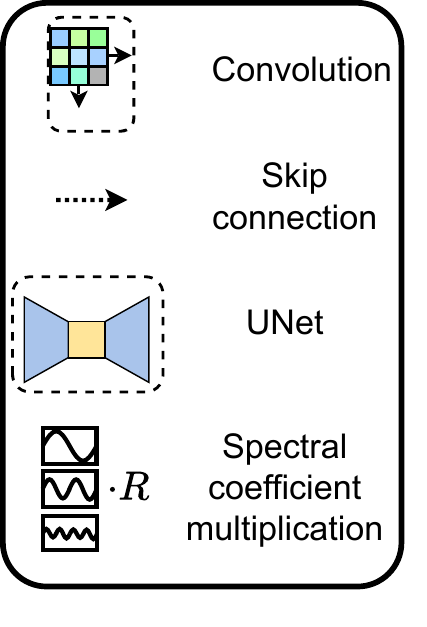}
\end{subfigure}
\caption{Schematic overview of model architectures.}
\label{fig:model-arch-diagrams}
\end{figure*}

\subsection{Model formulation}
We first introduce some notation and describe the general formulation along which we design the surrogate models. We denote the state of the system at time $t$ by $x^t$. Further, a sequence of $k$ states, spanning from sequence element $t$ up to but not including $t+k$ is denoted as $x^{t:t+k}$. %
Regarding the autoregressive approach, rather than using a model to predict a single timestep at a time, we formulate the model $f_\theta$ to operate on a \emph{bundle} of $k$ timesteps for both its input and output:
\begin{equation}
    \hat{x}^{t:t+k} = f_\theta(x^{t-k:t}).
\end{equation}
Notably, operating on the bundle level as opposed to single timesteps has been shown to improve performance, both in terms of computational efficiency as well as accuracy~\cite{brandstetter2022message}.

\subsection{Model designs}\label{subsec:model-designs}

We structure the model design space along the Encode-Process-Decode framework~\cite{sanchez-gonzalez2020}. Here, the task of the encoder is to map the input bundle $x^{t-k:t}$ to an abstract embedding $h^t$, representing the entire bundle $x^{t-k:t}$. The processor then operates on $h^t$ to produce an output embedding $o^t$. Finally, the decoder maps $o^t$ to a prediction for the next bundle $\hat{x}^{t:t+k}$. The architectural choices for the encoder, processor, and decoder components that we consider in this work are shortly described below.

\paragraph{Encoder.} As encoder we consider a simple convolutional neural network with a 1-by-1 kernel. Specifically, the different timepoints in the bundle are first flattened and taken as channels for the convolutional layer. Then, two convolutional layers are applied that operate on single grid points. 

\paragraph{Processor.} For the processor module, we consider three popular neural PDE surrogate modeling architectures; schematic overviews can be found in Figure~\ref{fig:model-arch-diagrams}. The first architecture is the Dilated ResNet (DRN)~\cite{stachenfeld2021learned}, illustrated in Figure~\ref{subfig:DRN-diagram}. The DRN is a convolutional model which effectively integrates information at different spatial scales by varying the dilation of subsequent convolutional layers: with larger dilations, long-range interactions can be modeled in a parameter-efficient manner. 

The second architecture is the U-FNO~\citep{wen2022ufno}, which is based on the Fourier Neural Operator (FNO)~\citep{li2021fno}; see Figure~\ref{subfig:FNO-diagram}. The FNO primarily relies on transforming the spatial signal to a frequency representation using the Fast Fourier Transform (FFT), and subsequently multiplying the spectral coefficients with a learned weight matrix $R$. Notably, this operation is equivalent to a global convolution, enabling the FNO to model long-range interactions. The U-FNO adds another parallel UNet branch, enabling it to better model high-frequency information, which is prominent in our setting near droplet boundaries. We either use only U-FNO layers or alternate them with regular FNO layers, as proposed in~\citet{wen2022ufno}.

The third architecture we consider is the UNet~\cite{ronneberger2015}, see Figure~\ref{subfig:UNet-diagram}. 
The UNet processes the input in two phases. In the downsampling phase, convolutional layers and strided convolutions are applied to downsample the grid to coarser representations. In the upsampling phase, transposed convolutions upsample the coarse grids to finer resolutions. These are concatenated with the outputs from the downsampling phase and processed by convolutional layers. 
We consider a UNet variant including residual blocks and %
normalization layers, which has achieved state-of-the-art performance in neural PDE surrogate modeling~\cite{gupta2022pdearena, lippe2023pderefiner}.

\paragraph{Decoder.} 
As decoder we consider a temporal convolutional model, which first reshapes $o^t$ appropriately and then applies a learned convolution along the temporal axis. This structure has shown to improve predictions by smoothing the output signal across the bundle~\cite{brandstetter2022message}.

\subsection{Adaptations for two-phase flow problems.}
While the aforementioned architectures have shown great results on various datasets, typical problems like turbulence and climate modeling do not necessarily share the same characteristics as two-phase flow problems. 
To tailor the methods to this specific task, we introduce several modifications that are compatible with all model architectures.

First, we condition the models on the geometry of the system. At each autoregressive step the field describing the boundaries -- the obstacles and the pore -- is concatenated along the channel axis, both for the encoder input and for the hidden layers in the processor. For the latter it is resampled to match the hidden representations' spatial grid if necessary, which only concerns UNet among the considered architectures. Additionally, after each step, the output is post-processed to zero out the non-fluid areas. This follows a similar approach to channel-based conditioning techniques for scalar PDE parameters, for example as in~\citet{takamoto2023cape}, but additionally preserves the spatial structure of the conditioning signal internally throughout all processor layers.

Second, all of the architectures are adapted to respect the horizontal periodic boundary conditions of the problem setting. For the convolutional models, adding appropriate circular padding takes care of this property. For the spectral component of the U-FNO, the FFT already respects periodicity, requiring no further adaptations.%

Third, we know that the mass of both phases should be preserved over time. However, in practice, the reference solver accumulates small numerical errors that slightly change the balance. To account for this we apply a correction after each prediction step, similar to~\citet{mcgreivy2023invariant}, but rather than enforcing exact mass conservation we only do so approximately. We apply a smooth clipping using a hyperbolic tangent, scaled according to the mass deviation and a maximum allowed deviation. For the total mass $m_{t+i}$ at future step $t+i$, using as reference the mass $m_t$ at last input timestep $t$, we renormalize as follows:
\begin{equation}
    \hat{m}_{t+i} = m_t \cdot \Big(1 + i\cdot\epsilon\tanh\Big(\frac{\nicefrac{m_{t+i}}{m_{t}}-1}{i \cdot\epsilon}\Big)\Big),
\end{equation}
where $\epsilon$ denotes the fraction of maximum mass deviation per timestep. Near perfect conservation results in almost no rescaling due do the linear behavior around 0, with a smooth clipping in the limits.

\section{Experiments}\label{sec:results}
\subsection{Data}
We construct a dataset consisting of simulations of water flowing over a dead-end pore filled with oil, similar to~\citet{linga2019bernaise}. We simulate up to physical time $T=50$. The spatial domain consists of a tube with a width of 3 and a height of 1, with a pore of a radius uniformly sampled from ${[0.05, 0.25]}$ at the top. Horizontal boundaries are periodic, effectively making it an infinite length tube. A surface charge is applied to induce electrowetting; charge values are sampled from the range ${[-10, -1]}$, biased towards more negative charges. The reasoning is that greater amplitudes result in more interesting behavior~\cite{linga2019bernaise}; at lower charges the droplet sometimes does not leave the pore.

To further increase the problem complexity we extend the numerical solver such that each simulation has one or two circular obstacles added to the internal domain. These circles have radii uniformly sampled from [0.05, 0.25] and have uniformly sampled center coordinates, while ensuring that they do not overlap, and taking into account a small minimum distance from the domain boundaries. We generate 920 simulations for this dataset, where 70\% is used for training, 10\% for validation, and 20\% for testing. In order to illustrate the comparative complexity, we also generate a baseline dataset with no obstacles, for which we generate 552 simulations with the same data split. For both datasets we model the phase field $\phi$, which describes the composition of the liquid as a value in ${[-1 ,1]}$: -1 denotes only water, 1 only oil, and other values indicate a mix of both.

The solver's temporal discretization $dt=0.02$ is sampled every 5 timesteps resulting in 500 timesteps per simulation. The finite element mesh with a resolution of 60 cells across the diameter is interpolated to an equidistant spatial grid of 96-by-64. The horizontal dimension is fixed, the vertical dimension is padded depending on the size of the pore, to unify all samples in the dataset to the same spatial dimension.

\subsection{Experimental setup}
The neural surrogate models $f_\theta$ learn to approximate a time-stepping operator acting on time bundles of the solution's phase field. Each bundle consists of 25 timesteps. Models are optimized using the MSE. To avoid overfitting on single-step predictions, we use pushforward training~\cite{brandstetter2022message}. Rather than unrolling once and computing the loss, we do multiple forward predictions of the model and compute the loss using the final prediction, where the loss is propagated only through the last prediction step. Starting at a single unrolling, the number of forward passes is increased by 1 each 25 epochs, up to a maximum of 8 (making the total unrolling window span 200 timesteps). We train for 500 epochs using the Adam optimizer~\cite{kingma2015} with an initial learning rate of $10^{-4}$, which is decayed by 0.4 at epochs 25, 125, 250 and 375. All models end with a hyperbolic tangent activation to map to the [-1, 1] phase field range. After each prediction we enforce boundary conditions within the domain and apply approximate mass conservation with a maximum deviation fraction of $\epsilon = 4 \cdot 10^{-4}$ per timestep. The data and code can be found on \url{https://github.com/yoeripoels/neural-pde-surrogates}.
\subsection{Results}
\begin{figure}[t]
\begin{subfigure}[t]{\linewidth}
\centering
\includegraphics[width=.9\linewidth]{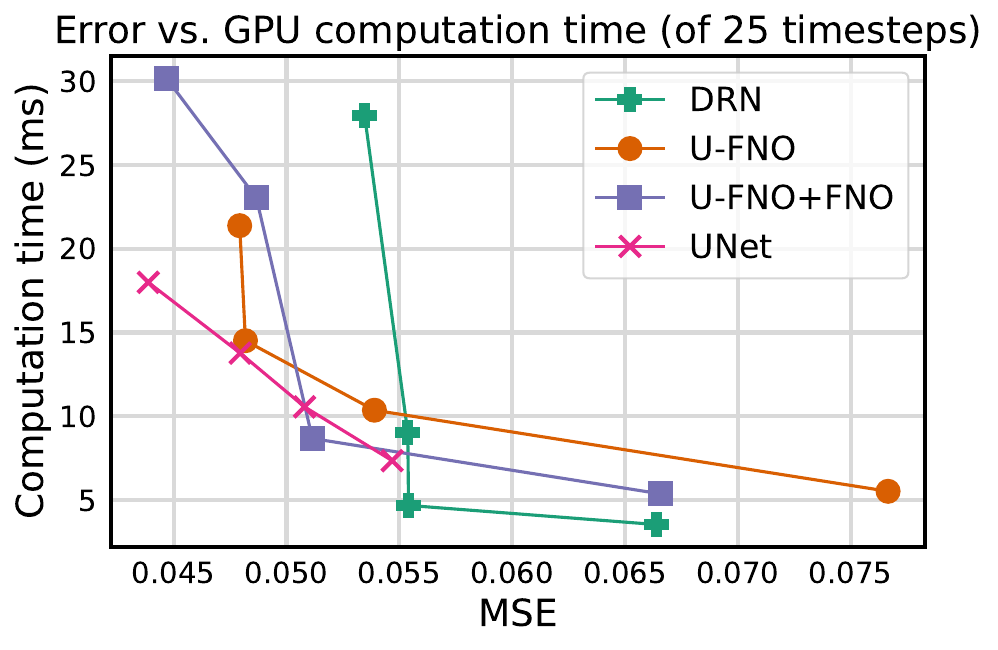}
\vspace{-0.1cm}
\caption{Results on the GPU (NVIDIA RTX 3080).}
\label{subfig:wpGPU}
\end{subfigure}
\begin{subfigure}[t]{\linewidth}
\centering
\hspace{-.12cm}\includegraphics[width=.91\linewidth]{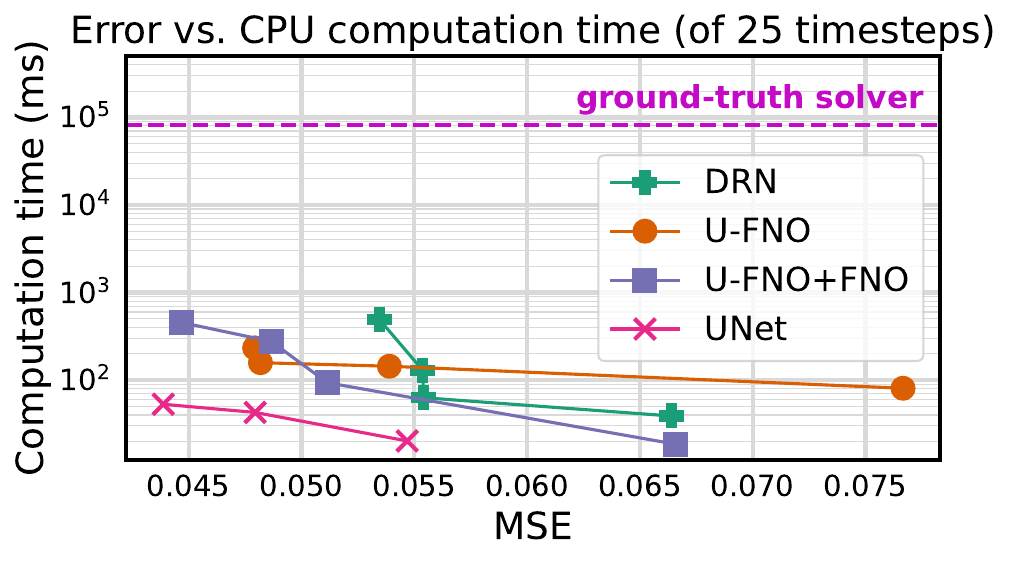}\hspace{10cm}
\vspace{-0.5cm}
\caption{Results on the CPU (AMD Ryzen 9 5900X), with the compute cost of the ground-truth solver as reference. Note that hyperparameter ranges were picked for GPU inference, and there may be settings yielding similar test errors better suited for CPU inference.}
\label{subfig:wpCPU}
\end{subfigure}
\caption{Work-precision diagrams for full rollout MSE on the test set versus the inference time of 25 timesteps (1 block), for the GPU and CPU. All models share the same general architecture with the processor parameters scaled. Parameter ranges were scanned to find ranges that scaled well w.r.t. both MSE and inference speed on the GPU; Pareto-optimal settings are plotted.}
\label{fig:wp}
\end{figure}

\begin{figure*}[t]
\hfill
\begin{subfigure}[t]{0.22\linewidth}
\centering
\vspace{-2.4cm}\includegraphics[width=\linewidth]{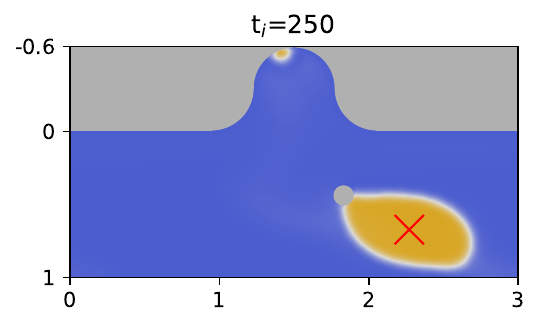}\hfill\vspace{0.1cm}
\caption{Visualization of a solution at $t=250$. The red `X' indicates the tracked center point.}
\label{subfig:diagrams1}
\end{subfigure}\hfill
\begin{subfigure}[t]{0.75\linewidth}
\centering
\includegraphics[width=.308\linewidth]{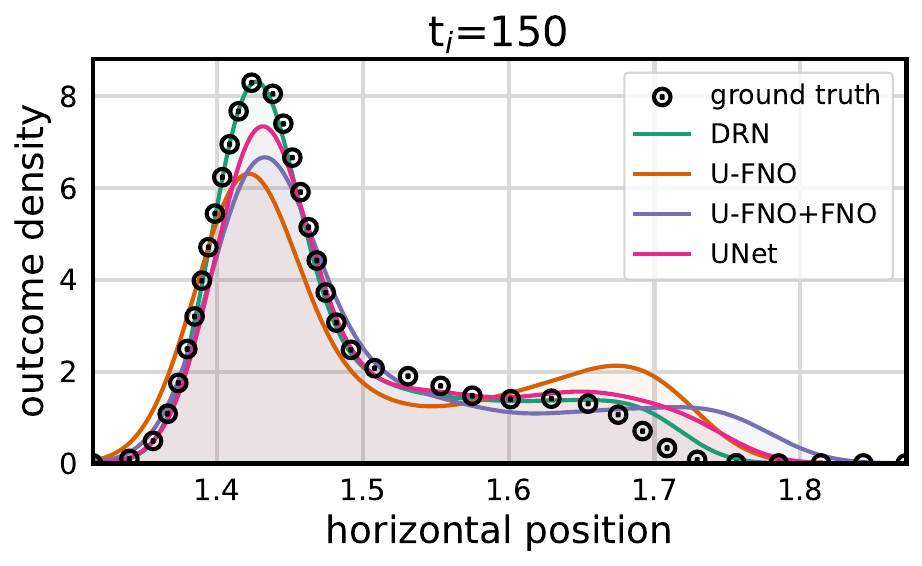}\hfill
\includegraphics[width=.308\linewidth]{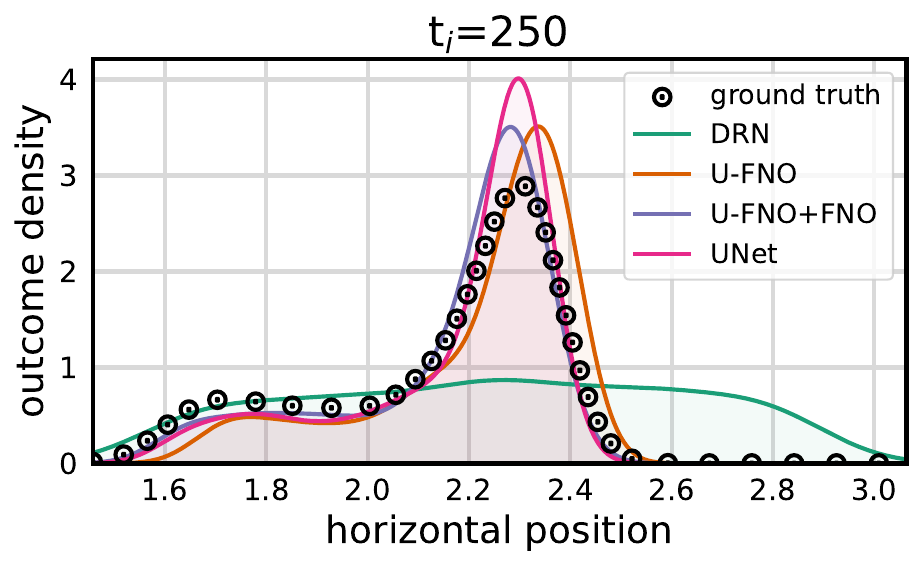}\hfill
\includegraphics[width=.324\linewidth]{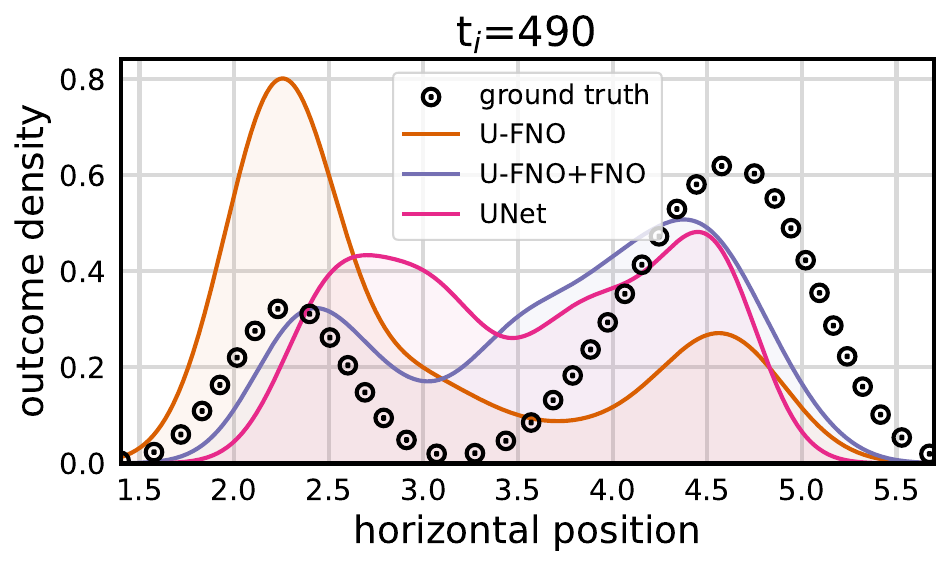}\hfill
\caption{Distributions of the droplet's horizontal position at various timesteps. A position exceeding the domain length (i.e. 3) indicates looping around due to the periodic BCs. The DRN is missing from the last plot due to the droplet having disintegrated at this timestep in multiple cases.}
\label{subfig:diagrams2}
\end{subfigure}
\caption{Comparison of distributions with small variations of the charge parameter, for a charge of -8.65 $\pm$ 20\% ($\pm$ 15 simulations). At earlier times the distributions mostly overlap, however in the final case there is a significant deviation. The multi-modal behavior -- where in some cases the droplet gets stuck on the obstacle, and in others it moves past -- is recovered in most cases, but not very precisely.}
\label{fig:sensitivity}
\end{figure*}

\textbf{Model comparison.} As main performance metrics we consider inference speed and the test-set MSE of the phase field, calculated over full simulation unrollings. Specifically, the initial block of $t=0$ up to $t=25$ is provided as input, after which the model autoregressively predicts $t=25$ to $t=500$. Computation time is measured as the inference time of 25 timesteps (1 block) using an NVIDIA RTX 3080 GPU and AMD Ryzen 9 5900X CPU. Models are implemented in PyTorch 2 and make use of \texttt{torch.compile} to speed up inference~\cite{ansel2024pytorch2}. At the time of writing, compilation for complex operators is not supported, making the computation time comparison somewhat unfavorable for U-FNO models.

We compare the different models and their respective speed-accuracy trade-offs in Figure~\ref{fig:wp}. All reported results are on the test set, and hyperparameters were selected such that they scale well with performance on the validation set; see Appendix~\ref{ap:params} for details. In general the UNet seems to provide the most favorable scaling. To place the computation time in context, we also plot the reference solvers cost for an equal timeframe, on equal hardware, in Figure~\ref{subfig:wpCPU}. We see that the neural surrogates can perform inference orders of magnitude faster than the numerical solver even on the same hardware. Futhermore, it is highly non-trivial to adapt existing numerical codes to exploit the parallel computation of GPUs, whereas neural surrogates can exploit this out-of-the-box. %
We note that this comparison needs to be viewed in the context of neural surrogates introducing some simulation error and requiring the generation of high-quality training data, as opposed to the numerical solver. Nonetheless, neural surrogates could enable many use cases, such as inverse design, where approximate solutions for a large search space of parameters that lie within the training distribution suffice. %

In addition to low MSE, in almost all cases the surrogates also produce simulations that qualitatively match well with the numerical solver. See for example Figure~\ref{fig:problem}, depicting a test set simulation alongside the best-performing UNet's predictions. For more qualitative results we refer to Appendix~\ref{ap:qual}.

\textbf{Parameter sensitivity.} As mentioned before, small changes in the input parameters can lead to significant changes in the spatiotemporal profile of the droplet behavior. An example of this bifurcating behavior occurs when, given certain charge parameters, the droplet gets stuck on an obstacle, but detaches after some time. This detachment time depends on the exact values of the electrical charge. In these cases, the pixel-wise MSE is a poor metric to evaluate whether the neural surrogate qualitatively matches the behavior of the numerical solver, as a small error in the predicted time of detachment would already lead to a large MSE. 

In Figure~\ref{fig:sensitivity} we evaluate such a case from the test set. We investigate the droplet movement for various perturbations of the surface charge parameter, with otherwise identical conditions. Each neural surrogate model architecture is evaluated using the configuration with the lowest MSE of Figure~\ref{fig:wp}. To compare whether they qualitatively match the reference simulations, we track the droplet's center and plot the distribution of the horizontal coordinate at several points in time. Over smaller time windows the distribution clearly matches the one generated by the numerical solver for most models. However, near the end of the simulation, the bimodal distribution is only approximately recovered, suggesting that neural PDE surrogates struggle to model such bifurcating dynamics over long time horizons.

\textbf{Ablations.} Finally, we confirm the benefit of the adaptations made to each of the model architectures. Table~\ref{tab:ablations} denotes these results: for each model, we take the best architectures (column `All') and disable the made adaptations. Column `$\setminus{\text{Inv}}$' denotes models with the invariances one can reasonably learn from data disabled. That is, these models do not have periodic boundary conditions enforced through the architecture, and do not have any mass conservation built in. For all models adding periodicity and mass conservation improved model performance. Column `$\setminus{\text{BC}}$' denotes models with no enforced spatial boundary conditions within the domain, that is, obstacles within the domain are not explicitly passed as conditioning in the internal model layers, and are not corrected after each autoregressive step. In most cases disabling this adaptation led to significant performance degradation.
\begin{table}[t]
\centering
\caption{Evaluations using the settings with the lowest errors. \textit{Middle:} Ablations of the introduced adaptations. $\setminus{\text{Inv}}$ denotes the enforced (approximate) invariances disabled, that is, no enforced periodicity and approximate mass conservation. $\setminus{\text{BC}}$ denotes no enforcement of the spatial boundary conditions of the geometry. \textit{Right:} Results when applying the model to the dataset with no obstacles, denoted as $\setminus{\text{Obs}}$. Even with no hyperparameter optimization, these models have considerably lower errors, indicating the challenging nature of the more complex setting we consider.}\vspace{0.3cm}
\begin{tabular}{l|ccc|c}

Model & All & $\setminus{\text{Inv}}$ & $\setminus{\text{BC}}$ & $\setminus{\text{Obs}}$ \\
\hline
DRN & \textbf{0.0535} & 0.0647 & 0.362 & 0.0168 \\
U-FNO & \textbf{0.0479} & 0.0513 & 0.395 & 0.0186 \\
U-FNO+FNO & \textbf{0.0447} & 0.0573 & 0.059 & 0.0167 \\
UNet & \textbf{0.0439} & 0.0465 & 0.112 & 0.0199 \\
\toprule
\end{tabular}

\label{tab:ablations}
\end{table}
Finally, to confirm the relative complexity of the adaptations that were made to the original oil expulsion setting, we benchmark all models on a dataset with no obstacles, as in~\citet{linga2019bernaise}. The results are shown in column `$\setminus{\text{Obs}}$'. Besides the absence of obstacles, the domains and simulation conditions are identical, making the error metrics lie on the same scale. The no-obstacle dataset contains approximately half of the number of training set simulations, and model hyperparameters were not optimized for this setting. Nevertheless, the test errors are two to three times lower, indicating that the dynamics get significantly more challenging by varying the geometry.

\section{Conclusion}\label{sec:conclusion}
We investigated neural PDE solvers for two-phase flows, more precisely for the expulsion of oil droplets using electrowetting. We extended the existing simulation setting of~\citet{linga2019bernaise} to more complex scenarios by introducing varying complex geometries in the domain. Here, the computational cost of for example inverse design tasks becomes prohibitively expensive using high-fidelity forward solves, making fast surrogate modeling approaches crucial. Further, the setting forms a harder and consequently more interesting benchmark for neural surrogate approaches. We investigated and extended three prevalent neural PDE solver methods in terms of their speed-accuracy trade-off, where we found the UNet to scale most favorably. Further, the UNet and U-FNO models both showed comparatively good qualitative performance when comparing aggregate properties of the droplet dynamics. Still, all models struggled to preserve these properties over long simulation horizons. Lastly, we showed the benefits of the introduced model extensions for the oil expulsion setting.

A future research direction would be to investigate more two-phase flow datasets~\cite{linga2019bernaise,JCPlinga, ALMASI2021} and to evaluate more neural PDE solver architectures, for example a selection of~\citet{lu2021learning, Cao2021Transformer, brandstetter2022message, tran2023factorized, li2023transformer, hao2023gnot, vcnef2024, serrano2023operator}. Additionally, a probabilistic approach could help in taking the uncertainties in bifurcating dynamics into account~\cite{cachay2023dyffusion, yang2023denoising, minartz2023equivariant, bergamin2024guided} or to mitigate autoregressive error accumulation~\cite{lippe2023pderefiner}. Finally, one could investigate the speed-up compared to reference numerical solvers and their scalability in more detail. In particular, while inference of the resulting networks is significantly faster, a dataset of expensive simulations must be generated up front. An interesting question is to consider the amortized cost of using neural surrogates in a downstream application. For example, one could do a large number of inverse problem optimizations using either only the high-fidelity numerical method, or by first generating a dataset, training a neural surrogate, and then consequently using a combination of both for the optimization. Such an evaluation could better evaluate the real-world applicability of the investigated methods.

\section{Acknowledgements}\label{sec:acknowledgements}
This work made use of the Dutch national e-infrastructure with the support of the SURF Cooperative using grant no. EINF-7709 and grant no. EINF-7724. The research of HB has received funding from the ERC under the European Union’s Horizon 2020 Research and Innovation Programme — Grant Agreement ID 818473.

\bibliography{references}
\bibliographystyle{icml2024}

\newpage
\appendix
\onecolumn
\section{Hyperparameters and results}\label{ap:params}
This appendix contains the hyperparameters and exact figures for runtime and accuracy for models that lie on the pareto fronts shown in Figure~\ref{subfig:wpGPU}. For UNets, unless otherwise mentioned, we use a fixed kernel size of 3 and a residual block depth of 1. For the DRN, the per-block dilation rates are fixed at (1, 2, 4, 8, 4, 2, 1). We refer to the respective papers~\cite{stachenfeld2021learned,gupta2022pdearena,wen2022ufno} and our code for more details.

\newcolumntype{P}[1]{>{\centering\arraybackslash}p{#1}}
\begin{table}[h]
\centering
\footnotesize
\caption{Pareto-optimal hyperparameters and results for DRN.}\label{tab:app-hyperparams-drn}
\begin{tabular}{@{}cccccc@{}}
\toprule
Layers & Kernel size & Hidden features & MSE   & GPU runtime (ms) & CPU runtime (ms) \\ \midrule
2      & 5           & 384                         & 0.053 & 27.9 & 488.8            \\
2      & 5           & 192                         & 0.055 & 9.0  &  129.4          \\
2      & 5           & 128                         & 0.055 & 4.7  &  63.1          \\
4      & 5           & 64                          & 0.066 & 3.5  &  39.0          \\ \bottomrule
\end{tabular}
\end{table}
\begin{table}[h]
\centering
\footnotesize
\caption{Pareto-optimal hyperparameters and results for U-FNO.}\label{tab:app-hyperparams-ufno}
\begin{tabular}{@{}ccccccc@{}}
\toprule
Layers & FNO modes & Hidden features & UNet channel multipliers & MSE   & GPU runtime (ms) & CPU runtime (ms) \\ \midrule
3      & 10        & 192             & (1,1)                    & 0.048 & 21.4   & 233.7         \\
2      & 10        & 192             & (1,1)                    & 0.048 & 14.5  &  158.3         \\
2      & 10        & 192             & (1)                     & 0.054 & 10.4   &  144.2        \\
1      & 10        & 192             & (1, 1)                      & 0.077 & 5.5  & 80.8           \\ \bottomrule
\end{tabular}
\end{table}
\begin{table}[H]
\centering
\footnotesize
\caption{Pareto-optimal hyperparameters and results for U-FNO + FNO.}\label{tab:app-hyperparams-ufno-fno}
\begin{tabular}{@{}ccccccc@{}}
\toprule
Layers                & FNO modes & Hidden features & UNet channel multipliers & MSE   & GPU runtime (ms) & CPU runtime (ms) \\ \midrule
(FNO, U-FNO) $\times 2$ & 10        & 256             & (1,1,1,1)                & 0.045 & 30.2   & 455.3         \\
(FNO, U-FNO) $\times 2$ & 10        & 256             & (1,1)                    & 0.049 & 23.0  & 276.3          \\
(FNO, U-FNO)            & 10        & 192             & (1,1)                    & 0.051 & 8.7 &  93.5           \\
(FNO, U-FNO)            & 6         & 64              & (1)                      & 0.067 & 5.4 &  18.5           \\ \bottomrule
\end{tabular}
\end{table}

\begin{table}[h]
\centering
\footnotesize
\caption{Pareto-optimal hyperparameters and results for UNet.}\label{tab:app-hyperparams-unet}
\begin{tabular}{@{}ccccccc@{}}
\toprule
 & Channel multipliers & Residual block depth & Hidden features (1st layer) & MSE    & GPU runtime (ms) & CPU runtime (ms) \\ \midrule
 & (2,2,1,2)                  & 2                    & 32                          & 0.044 & 18.0 & 53.0             \\
 & (2,2,2)    & 2                    & 32                          & 0.048 & 13.8  & 42.8           \\
 & (2,2)               & 2                    & 64                          & 0.051 & 10.6 & 66.0            \\
 & (2,2)            & 1                    & 32                          & 0.055 & 7.4 & 20.2             \\ \bottomrule
\end{tabular}
\end{table}
\newpage

\section{Additional qualitative results}\label{ap:qual}
This appendix contains additional qualitative results for all model architectures. For each model architecture we use the hyperparameters that resulted in the lowest MSE. We plot three examples: a case that works well for all in Figure~\ref{subfig:A}, a mixed case in Figure~\ref{subfig:B}, and a failure case in Figure~\ref{subfig:C}. In the last, one can clearly see that the models struggle with the outlier case of an even split of the droplet into two separate droplets.

\begin{figure*}[h]
\hfill
\begin{subfigure}[t]{\linewidth}
\raisebox{6.59cm}{\rotatebox{90}{\textbf{Reference}}}\hspace{0.13cm}
\includegraphics[width=.18\linewidth]{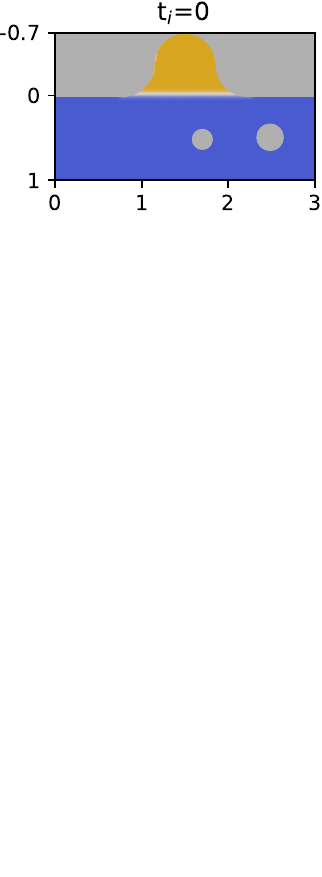}\hfill
\hspace{-0.33cm}\raisebox{0.6cm}{\rotatebox{90}{\hphantom{x\textbf{UNet}}\hphantom{xxx}\textbf{U-FNO}}}\hspace{0.1cm}\raisebox{0.6cm}{\rotatebox{90}{\hphantom{.}\textbf{UNet}\hphantom{.xxx.}\textbf{+FNO}\hphantom{xxx.}\textbf{U-FNO}\hphantom{xxx.}\textbf{DRN}}}\hspace{0.03cm}
\includegraphics[width=.18\linewidth]{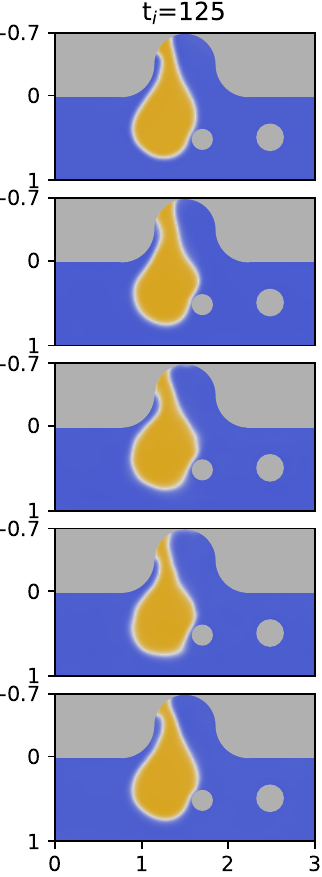}\hfill
\includegraphics[width=.18\linewidth]{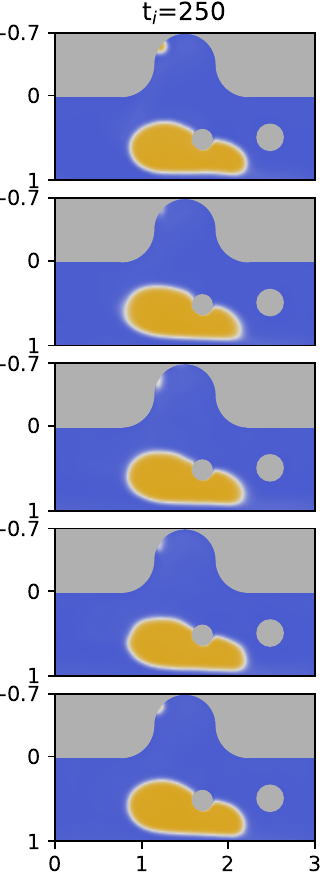}\hfill
\includegraphics[width=.18\linewidth]{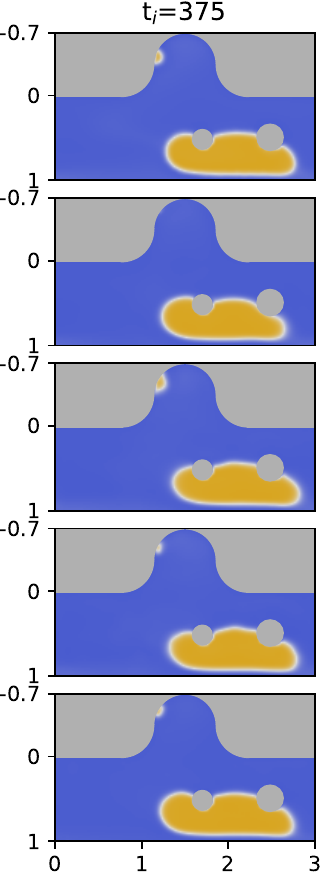}\hfill
\includegraphics[width=.18\linewidth]{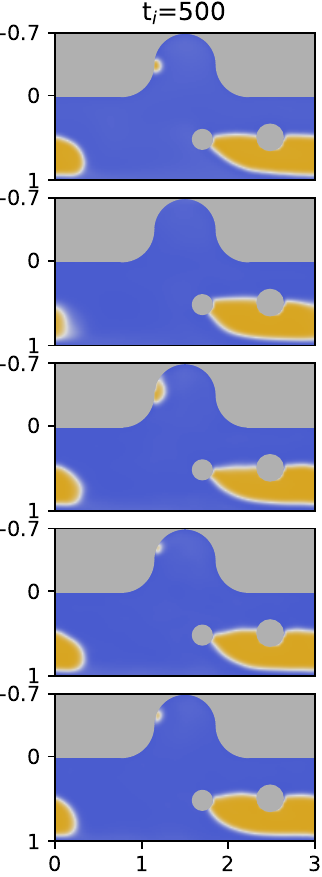}\hfill
\hfill
\caption{Example with two obstacles; all models qualitatively match the reference solution.}\label{subfig:A}
\end{subfigure}
\begin{subfigure}[b]{\linewidth}
\raisebox{6.07cm}{\rotatebox{90}{\textbf{Reference}}}\hspace{0.13cm}
\includegraphics[width=.18\linewidth]{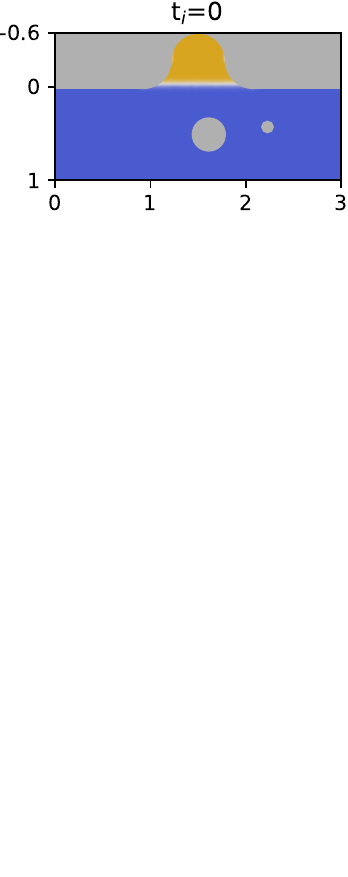}\hfill
\hspace{-0.33cm}\raisebox{0.4cm}{\rotatebox{90}{\hphantom{x\textbf{UNet}}\hphantom{xxx}\textbf{U-FNO}}}\hspace{0.1cm}\raisebox{0.4cm}{\rotatebox{90}{\hphantom{x}\textbf{UNet}\hphantom{xxx.}\textbf{+FNO}\hphantom{xx.}\textbf{U-FNO}\hphantom{xxx}\textbf{DRN}}}\hspace{0.03cm}
\includegraphics[width=.18\linewidth]{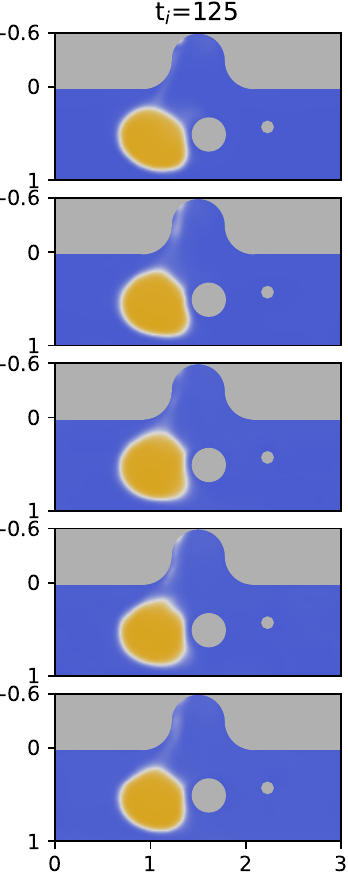}\hfill
\includegraphics[width=.18\linewidth]{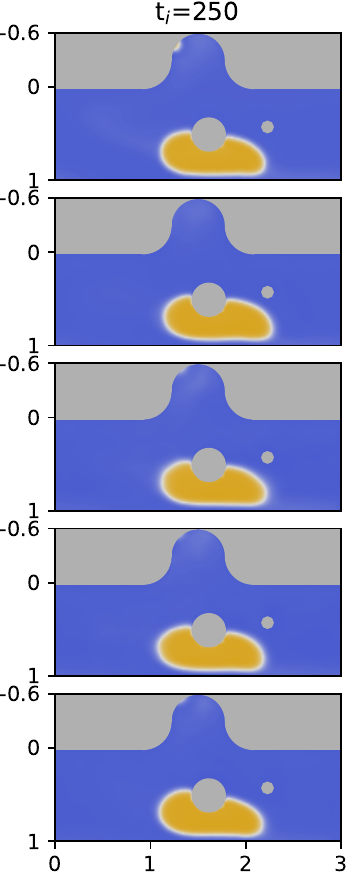}\hfill
\includegraphics[width=.18\linewidth]{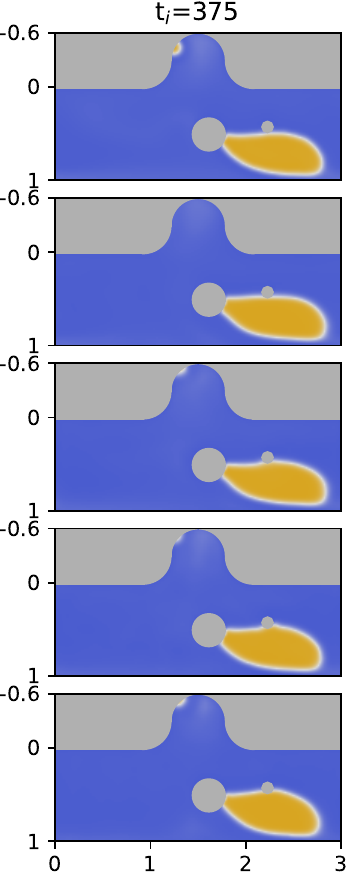}\hfill
\includegraphics[width=.18\linewidth]{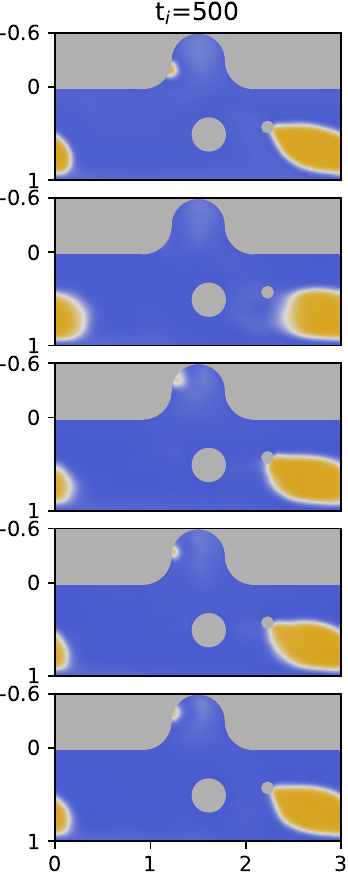}\hfill
\hfill
\caption{Example with two obstacles; here, the DRN struggles to reproduce the behavior near the final timestep.}\label{subfig:B}
\end{subfigure}
\end{figure*}
\begin{figure*}\ContinuedFloat
\hfill
\begin{subfigure}[t]{\linewidth}
\raisebox{7.2cm}{\rotatebox{90}{\textbf{Reference}}}\hspace{0.13cm}
\includegraphics[width=.18\linewidth]{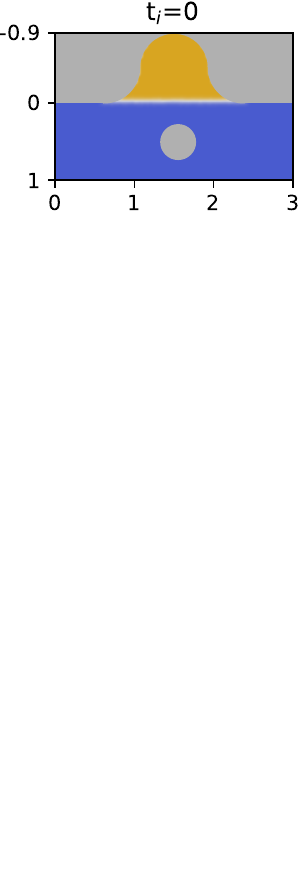}\hfill
\hspace{-0.33cm}\raisebox{0.6cm}{\rotatebox{90}{\hphantom{x\textbf{UNet}}\hphantom{xxxx}\textbf{U-FNO}}}\hspace{0.1cm}\raisebox{0.6cm}{\rotatebox{90}{\hphantom{x}\textbf{UNet}\hphantom{xxxx.}\textbf{+FNO}\hphantom{xxx..}\textbf{U-FNO}\hphantom{xxxx}\textbf{DRN}}}\hspace{0.03cm}
\includegraphics[width=.18\linewidth]{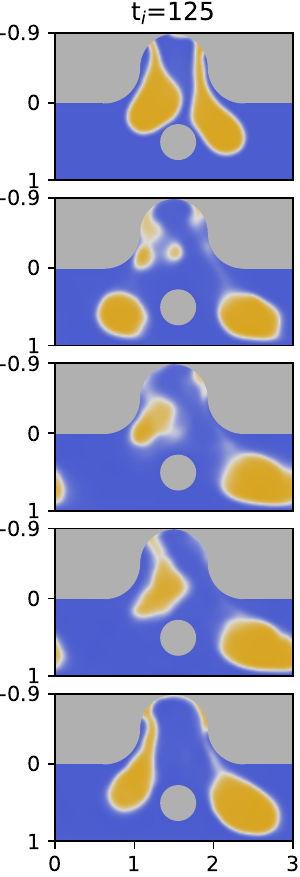}\hfill
\includegraphics[width=.18\linewidth]{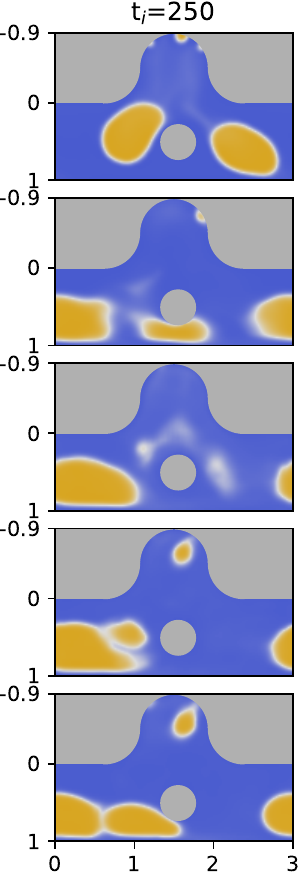}\hfill
\includegraphics[width=.18\linewidth]{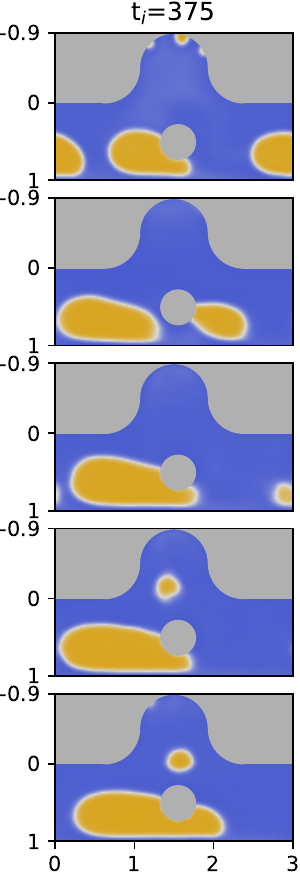}\hfill
\includegraphics[width=.18\linewidth]{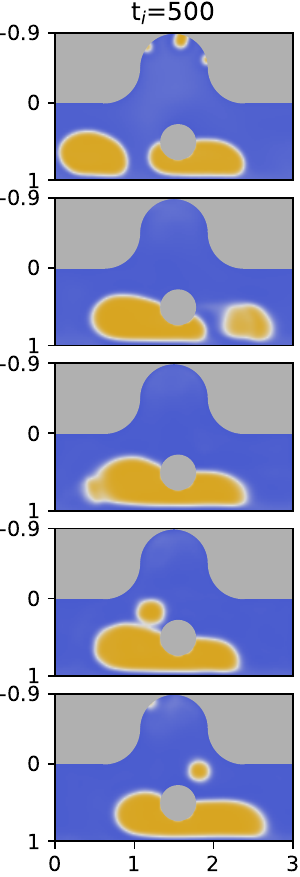}\hfill
\hfill
\caption{Failure mode for the neural PDE surrogates: they cannot accurately reproduce the outlier case of an evenly splitting droplet. At some point during the simulation all models merge the two droplets back into one, which does not happen in the reference solution.}\label{subfig:C}
\end{subfigure}
\caption{Test set example with predictions of all model classes considered, using the settings within each class that led to the lowest MSE.}
\label{fig:qualitative-appendix}
\end{figure*}
\vspace{100cm}\phantom{x} %

\end{document}